\documentclass[conference]{IEEEtran}
\IEEEoverridecommandlockouts

\usepackage{cite}
\usepackage{amsmath,amssymb,amsfonts}
\usepackage{algorithmic}
\usepackage{graphicx}
\usepackage{textcomp}
\usepackage[table]{xcolor}
\usepackage{booktabs}
\usepackage{multirow}
\usepackage{url}

\def\BibTeX{{\rm B\kern-.05em{\sc i\kern-.025em b}\kern-.08em
    T\kern-.1667em\lower.7ex\hbox{E}\kern-.125emX}}

\makeatletter
\renewcommand{\footnoterule}{
  \kern -3pt
  \hrule width 0.33\columnwidth height 0.4pt
  \kern 2.6pt
}
\makeatother

\begin{document}

\title{Context-structured Video Anomaly Detection with Large Vision-Language Models
}

\author{ \IEEEauthorblockN{ Dongjun Kim\textsuperscript{1} \quad Changjae Oh\textsuperscript{2} \quad Andrea Cavallaro\textsuperscript{3} \quad Jeonghoon Mo\textsuperscript{1,*} } \IEEEauthorblockA{ \textsuperscript{1}Yonsei University, Seoul, Republic of Korea\\ \textsuperscript{2}Queen Mary University of London, London, United Kingdom\\ \textsuperscript{3}EPFL, Lausanne, Switzerland\\[0.3em] {\footnotesize\ttfamily \url{https://dj991108.github.io/CSI-VAD/}} } \thanks{\textsuperscript{*}Corresponding author: j.mo@yonsei.ac.kr.} }

\maketitle

\begin{abstract}
Training video anomaly detectors is challenging due to the difficulty and cost of annotating diverse and rare abnormal events. Although recent large vision-language models enable training-free inference, existing approaches mostly rely on holistic inference over sampled video and may miss context-specific anomaly cues. In this paper, we present CSI-VAD, a training-free video anomaly detector that identifies abnormal events across diverse contexts. The key idea is to decompose each video into three distinct contexts (environment, objects, time) and perform context-specific inference in separate branches. Because we ground anomaly judgments solely in context-specific visual cues, we do not require predefined text prompts describing abnormal events or dataset-specific tuning. Experiments on UCF-Crime and UBnormal show that CSI-VAD consistently improves over the direct holistic baseline and achieves competitive performance against existing methods, showing the advantage of structured context decomposition for training-free video anomaly detection.
\end{abstract}


\section{Introduction}
\label{sec:intro}

Video anomaly detection (VAD) methods for long, untrimmed videos are mostly training-based, relying on large-scale annotated anomaly data with carefully designed training recipes~\cite{sultani2018real,tian2021weakly,chen2023mgfn}. Real-world anomalies are rare and diverse~\cite{sultani2018real,tian2021weakly} and performance can degrade under viewpoint shift, which require additional data collection and re-training. These challenges motivate a training-free alternative that can generalize to unseen environments without specific fine-tuning.

The large-scale pre-training of large vision-language models (LVLMs) provides visual and linguistic priors that transfer without task-specific anomaly training~\cite{alayrac2022flamingo,li2023blip2}. Recent VAD approaches leverage LVLMs to infer abnormality from sampled video frames using general visual and linguistic knowledge~\cite{chen2023tevad,zanella2024harnessing}.
In many LVLM-based VAD pipelines, sampled frames from the entire video are processed in a single end-to-end inference for a global abnormal/normal decision~\cite{zanella2024harnessing}.
However, existing approaches largely rely on holistic video-wide inference, while abnormal events often emerge across multiple \emph{local contexts}, such as environmental conditions, object behaviors and interactions, and event dynamics over time~\cite{pang2021deep,yang2024context}.
This global inference forces the model to simultaneously capture, interpret and integrate multiple contextual perspectives, thus missing context-specific anomaly cues.

To make use of context-specific anomaly cues, we propose {C}ontext-{S}tructured {I}nference for {V}ideo {A}nomaly {D}etection (CSI-VAD), a training-free VAD framework that reduces inference complexity by decomposing holistic video-level anomaly judgment into complementary branches: (i) \textit{environment context}, for describing the spatial scene conditions, (ii)
\textit{object context}, for consisting of object-centric overlay frames with detection and tracking; and (iii) \textit{time context}, for summarizing event dynamics across the video.
Each context is analyzed independently via dedicated recognition modules by an LVLM/LLM that produces a binary label, an anomaly score and a brief explanation. We then aggregate context-level outputs using ensemble and voting rules to produce the video-level decision.
The anomaly judgment is grounded solely on context-specific visual perception cues and, therefore, CSI-VAD does not require predefined text prompts describing abnormal events or dataset-specific tuning.

\begin{figure*}[!t]
    \centering
    \includegraphics[width=\textwidth]{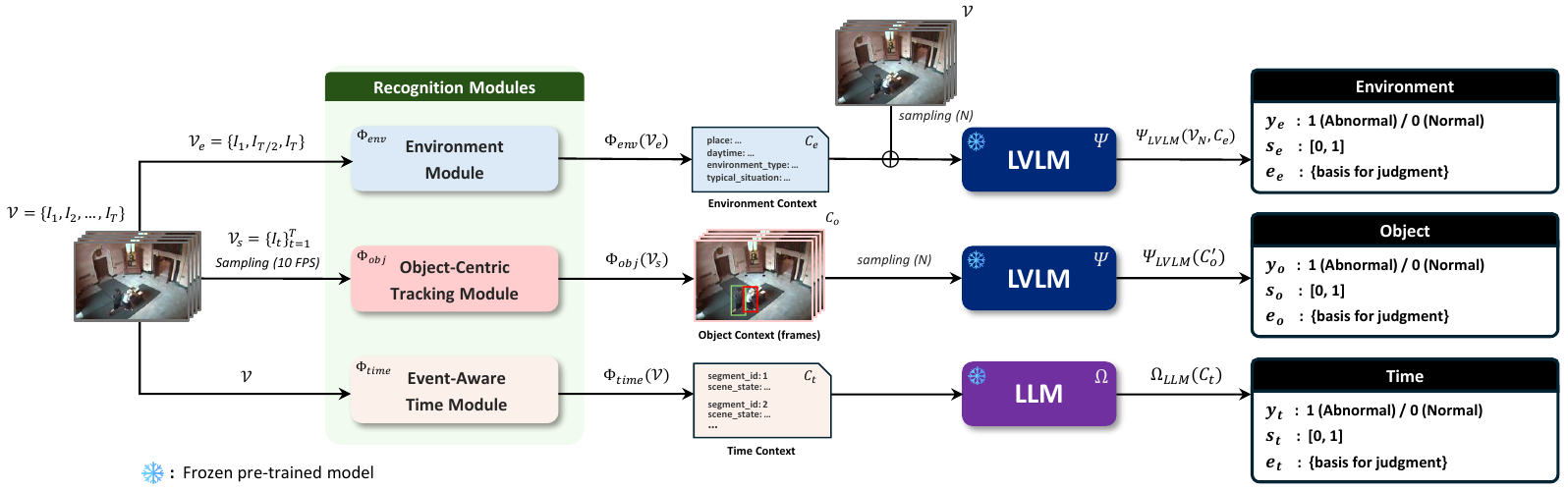}
    
    \caption{\textbf{Overview of the proposed CSI-VAD framework.} We extract environment ($C_e$), object ($C_o$), and time ($C_t$) contexts from an input video $\mathcal{V}$. Each context is then independently processed by an LVLM/LLM to produce a label $y_{(\cdot)}$, anomaly score $s_{(\cdot)}$, and explanation $e_{(\cdot)}$, and the outputs are combined for the final video-level prediction.}
    
    \label{fig:framework}
\end{figure*}

Our main contributions are:
\begin{itemize}
    \item a context-structured training-free VAD framework that replaces single-step global inference with decomposed inference over environment, object and time contexts, thereby reducing inference complexity;

    \item dedicated recognition modules that operationalize this decomposition by constructing distinct context representations, including scene-level environment priors, object-centric tracking overlays, and event-aware summaries over time, allowing each branch to focus on a separate source of anomaly evidence;

    \item score- and label-level aggregation strategies that leverage complementary context evidence for different decision needs: score aggregation enables flexible anomaly ranking by integrating context-wise anomaly strengths, while label aggregation controls sensitivity and strictness trade-offs through an adjustable decision threshold.
\end{itemize}

We validate CSI-VAD on UCF-Crime~\cite{sultani2018real} and UBnormal~\cite{acsintoae2022ubnormal} under video-level evaluation, showing consistent improvements over the direct holistic baseline and competitive performance against existing VAD methods.

\section{Method}
\label{sec:method}

Given an input video $\mathcal{V}=\{I_1,\dots,I_T\}$ with $T$ frames, we construct complementary context representations and perform anomaly inference for each context (see Fig.~\ref{fig:framework}). 
The recognition modules extract three context representations from $\mathcal{V}$: an environment context $C_e$, an object context $C_o$, and a time context $C_t$ (Sec.~\ref{subsec:modules}).
The corresponding context-wise outputs are subsequently aggregated to obtain the final video-level prediction (Sec.~\ref{subsec:aggregation}).

\subsection{Recognition modules}
\label{subsec:modules}

We uniformly sample $N$ frames from the original video and from the object context to form $\mathcal{V}_N$ and $C'_o$, respectively.
The time branch directly uses $C_t$ as input to an LLM.
Each context is processed independently using dedicated LVLM/LLM modules and produces an anomaly judgment, i.e. a binary label $y_{\textrm(\cdot)}$, an anomaly score $s_{\textrm(\cdot)}\in[0,1]$, and a linguistic explanation $e_{\textrm(\cdot)}$:
\begin{equation}
\label{eq:csi_framework}
\begin{aligned}
(y_e, s_e, e_e) &= \Psi_{\text{LVLM}}\!\left(\mathcal{V}_N, C_e\right),\\
(y_o, s_o, e_o) &= \Psi_{\text{LVLM}}\!\left(C'_o\right),\\
(y_t, s_t, e_t) &= \Omega_{\text{LLM}}\!\left(C_t\right).
\end{aligned}
\end{equation}

\begin{figure*}[t]
    \centering
    \includegraphics[width=\textwidth]{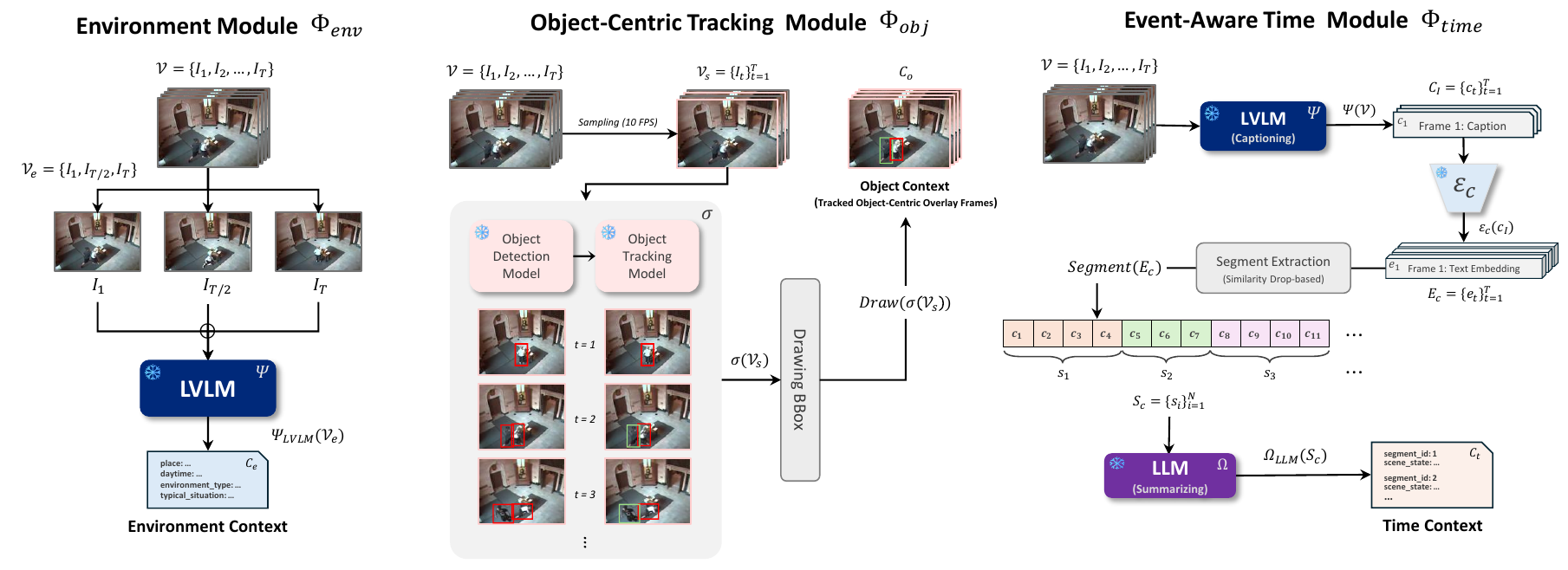}
    \caption{\textbf{Recognition modules}. The framework captures environment, object, and time contexts via three dedicated modules $\Phi_{(\cdot)}$.}
    \label{fig:recognition_modules}
\end{figure*}

In Fig.~\ref{fig:recognition_modules}, we show how each module is designed to extract the context-specific representations.

\noindent {\bf{Environment module}} ($\Phi_{\text{env}}$). We construct a scene-level context $C_e$, capturing the spatial conditions and the typical situation implied by the scene.
Given $\mathcal{V}$, we select three keyframes $\mathcal{V}_e=\{I_1, I_{\lfloor T/2 \rfloor}, I_T\}$ to represent the beginning, middle, and end of the video, and apply a multi-image LVLM to obtain $C_e=\Psi_{\text{LVLM}}(\mathcal{V}_e)$. The resulting $C_e$ includes \textit{place}, \textit{daytime}, \textit{environment type} (a brief description of the place), and a \textit{typical situation}, where the latter summarizes visible infrastructure and the activities normally afforded by the scene.
By explicitly encoding a scene prior from a small set of representative frames, the module provides contextual grounding for anomaly detection in the environment branch.

\noindent {\bf{Object-centric tracking module}} ($\Phi_{\text{obj}}$). We construct an object context $C_o$, an object-emphasized visual sequence designed to highlight object-level anomaly cues while preserving the original scene context.
From the input video $\mathcal{V}$, we first sample frames at 10 FPS to obtain a temporally dense sequence $\mathcal{V}_s=\{I_t\}_{t=1}^{T'}$.
An object detection model and an object tracking model are then applied to $\mathcal{V}_s$ to produce tracking results $\sigma(\mathcal{V}_s)$, which are rendered on the original sampled frames to generate the overlay-based object context $C_o=\mathrm{Draw}(\sigma(\mathcal{V}_s))$.
Different tracked instances are represented using different box colors, making object identity and time continuity visually explicit.
This representation preserves the scene appearance of the original video while emphasizing object locations, motion, and interactions, allowing the LVLM to focus on object-related anomaly evidence.
The resulting object context $C_o$ is uniformly sampled to form the object-branch input $C'_o$.

\noindent {\bf{Event-aware time module}} ($\Phi_{\text{time}}$). We capture the progression of events across the video.
Instead of performing inference over the full frame sequence directly, the module first generates frame-level captions from sparsely sampled frames using an LVLM, producing a temporally ordered caption sequence $C_I=\{c_t\}_{t=1}^{T}=\Psi(\mathcal{V})$.
These captions are converted into text embeddings $E_c=\{e_t\}_{t=1}^{T}$, and the sequence is partitioned into semantically coherent segments $\mathrm{Segment}(E_c)$ by detecting sustained drops in similarity.
The resulting caption groups $S_c=\{s_i\}_{i=1}^{N}$ are then summarized by an LLM to obtain the time context $C_t=\Omega_{\text{LLM}}(S_c)$, where each segment is represented by a concise \textit{scene state}.
By transforming a long video into an ordered sequence of semantically segmented and summarized units over time, the module enables the time branch to focus on high-level event flow and state transitions that may indicate abnormal activity.

\subsection{Aggregation strategy}
\label{subsec:aggregation}

We aggregate the outputs from $\Phi_{\text{env}}$, $\Phi_{\text{obj}}$, and $\Phi_{\text{time}}$ to obtain a final video-level decision. We separately consider score and label aggregation. Score aggregation combines the context-wise anomaly scores into a single video-level score used for ranking-based evaluation, whereas label aggregation combines binary context predictions via a voting rule for binary classification evaluation. The two strategies serve different evaluation objectives and are applied independently. This design allows the framework to integrate complementary evidence from environment, object, and time branches.

\noindent\textbf{Score aggregation.}
We combine the context-wise anomaly scores $s_e, s_o, s_t \in [0,1]$ using three alternative strategies that are treated as independent design variants:
\begin{subequations}
\label{eq:score_agg}
\begin{align}
\hat{s}_{\text{max}} &= \max \left(s_e, s_o, s_t\right), \label{eq:score_agg_max}\\
\hat{s}_{\text{mean}} &= \dfrac{s_e + s_o + s_t}{3}, \label{eq:score_agg_mean}\\
\hat{s}_{\text{nor}} &= 1 - (1-s_e)(1-s_o)(1-s_t), \label{eq:score_agg_nor}
\end{align}
\end{subequations}
where $\hat{s}_{\text{max}}$, $\hat{s}_{\text{mean}}$, and $\hat{s}_{\text{nor}}$ are the video-level scores obtained by Max Ensemble, Mean Ensemble, and Noisy-OR, respectively~\cite{kittler1998combining,pearl1988probabilistic}. Max Ensemble retains the strongest anomaly cue among the three contexts, Mean Ensemble reflects the overall anomaly tendency across contexts through equal averaging, and Noisy-OR increases the final score when multiple branches simultaneously provide positive anomaly evidence.

\noindent\textbf{Label aggregation.}
We combine the context-wise labels $y_e, y_o, y_t \in \{0,1\}$ for binary decision making using a threshold-based voting rule~\cite{lam1997majority} that enables controllable trade-offs between sensitivity and strictness in the final video-level prediction:
\begin{equation}
\label{eq:label_agg}
\hat{y} =
\begin{cases}
1, & \text{if } |\mathcal{C}| \ge N,\\
0, & \text{otherwise},
\end{cases}
\quad
\mathcal{C}=\{k \in \{e,o,t\}\mid y_k = 1\}.
\end{equation}
By varying $N$, the aggregation rule instantiates OR voting when $N=1$, majority voting when $N=2$, and AND voting when $N=3$. 

\section{Experiments}
\label{sec:experiments}
We evaluate CSI-VAD on two benchmarks, UCF-Crime \cite{sultani2018real} and UBnormal \cite{acsintoae2022ubnormal}, under a video-level setting. UCF-Crime uses a test split of 290 videos (140 abnormal and 150 normal) covering 13 abnormal event categories, while UBnormal contains 211 test videos (53 normal and 158 abnormal). For both benchmarks, each method outputs an anomaly score and a binary label for the entire video. We use AUC-ROC for score-based evaluation, and Accuracy, Precision, Recall, and F1 score for binary classification. Qwen2.5-VL-7B-Instruct \cite{Qwen2.5-VL} is used for the LVLM-based branches, and Qwen2.5-7B-Instruct \cite{qwen2.5} is used for the time inference branch. In the object branch, we employ RF-DETR (Medium) \cite{robinson2025rf} for object detection and StrongSORT \cite{du2023strongsort} for tracking. As the baseline, we use the same LVLM backbone to directly infer the video-level score and label from the input video without context decomposition. All experiments are conducted on a single NVIDIA A100 GPU with 80GB memory.

\subsection{Quantitative results}
\label{subsec:quantitative_results}

Table~\ref{tab:auc_existing} summarizes the AUC-ROC results of representative existing video-level VAD methods and our training-free variants (\textit{Ours}$_{256}$) on UCF-Crime under the 256-frame setting, which shows the best performance in our experiments.
The results of existing methods are taken from their original publications using their evaluation protocols. CSI-VAD achieves competitive performance in this comparison while other method are trained on UCF-Crime. In particular, \textit{Ours}$_{256}$ with Mean Ensemble and \textit{Ours}$_{256}$ with Noisy-OR outperform strong training-based methods such as DUAL VAD \cite{jung2025dual} and Anomaly Scorer \cite{gao2025stead} (Base, X3D). These results highlight that our context-structured inference framework is effective and demonstrates strong generalization without task-specific training.

\begin{table}[t]
\caption{Comparison with representative existing video-level anomaly detection methods on UCF-Crime. Results for prior methods are taken from their original publications and follow their respective evaluation protocols. The best result is shown in bold and the second-best is underlined.}
\label{tab:auc_existing}
\centering
\scriptsize
\setlength{\tabcolsep}{3.5pt}
\renewcommand{\arraystretch}{0.95}
\resizebox{\columnwidth}{!}{
\begin{tabular}{p{0.58\columnwidth}cc}
\toprule
\textbf{Method} & \textbf{Training-Free} & \textbf{AUC-ROC (\%)} \\
\midrule
Sultani et al. (I3D)~\cite{sultani2018real} & $\times$ & 77.92 \\
RTFM (I3D)~\cite{tian2021weakly} & $\times$ & 84.03 \\
S3R (I3D)~\cite{wu2022self} & $\times$ & 85.99 \\
MGFN (I3D)~\cite{chen2023mgfn} & $\times$ & 86.98 \\
PEL (I3D)~\cite{pu2024learning} & $\times$ & 86.76 \\
BN-WVAD (I3D)~\cite{zhou2024batchnorm} & $\times$ & 87.24 \\
Anomaly Scorer (Base, X3D)~\cite{gao2025stead} & $\times$ & 91.34 \\
DUAL VAD~\cite{jung2025dual} & $\times$ & 92.33 \\
\midrule
\textit{Ours}$_{256}$ (Max Ensemble) & $\checkmark$ & 89.33 \\
\textbf{\textit{Ours}$_{256}$ (Mean Ensemble)} & $\checkmark$ & \textbf{92.90} \\
\textbf{\textit{Ours}$_{256}$ (Noisy-OR)} & $\checkmark$ & \underline{92.88} \\
\bottomrule
\end{tabular}
}
\end{table}

In Table~\ref{tab:auc_recent_methods}, we further compare CSI-VAD with recent VAD methods, including both training-free (LAVAD \cite{zanella2024harnessing}) and training-based (VERA \cite{ye2025vera}, PEL4VAD \cite{pu2024learning}) methods, using UCF-Crime but with a different evaluation protocol. We follow the protocol in \cite{borodin2025benchmarking}, using 0.5 fps sampling for videos shorter than 60 seconds and 32-frame uniform sampling for videos of 60 seconds or longer. 
Under this protocol, CSI-VAD consistently outperforms LAVAD \cite{zanella2024harnessing} across all aggregation strategies, showing that the proposed context-structured inference provides a clear advantage over a competitive training-free baseline.
Moreover, despite being training-free, CSI-VAD also surpasses the training-based VERA \cite{ye2025vera} and remains competitive with PEL4VAD \cite{pu2024learning}, which achieves the highest AUC in this comparison. This result further supports the effectiveness of CSI-VAD across different evaluation settings.

\begin{table}[t]
\caption{Comparison with recent VAD methods on UCF-Crime under the evaluation protocol used in prior work: 0.5 fps sampling for videos shorter than 60 seconds and 32-frame uniform sampling for videos of 60 seconds or longer. The best result is shown in bold and the second-best is underlined.}
\label{tab:auc_recent_methods}
\centering
\scriptsize
\setlength{\tabcolsep}{3.5pt}
\renewcommand{\arraystretch}{0.95}
\resizebox{0.8\columnwidth}{!}{
\begin{tabular}{lcc}
\toprule
\textbf{Method} & \textbf{Training-Free} & \textbf{AUC-ROC (\%)} \\
\midrule
LAVAD~\cite{zanella2024harnessing} & $\checkmark$ & 77.19 \\
VERA~\cite{ye2025vera} & $\times$ & 64.60 \\
PEL4VAD~\cite{pu2024learning} & $\times$ & \textbf{87.31} \\
\midrule
Ours (Max Ensemble) & $\checkmark$ & 85.48 \\
Ours (Mean Ensemble) & $\checkmark$ & 85.96 \\
Ours (Noisy-OR) & $\checkmark$ & \underline{86.08} \\
\bottomrule
\end{tabular}
}
\end{table}

\subsection{Qualitative results}
\label{subsec:qualitative_results}

\begin{figure}[t]
    \centering
    \includegraphics[width=\columnwidth]{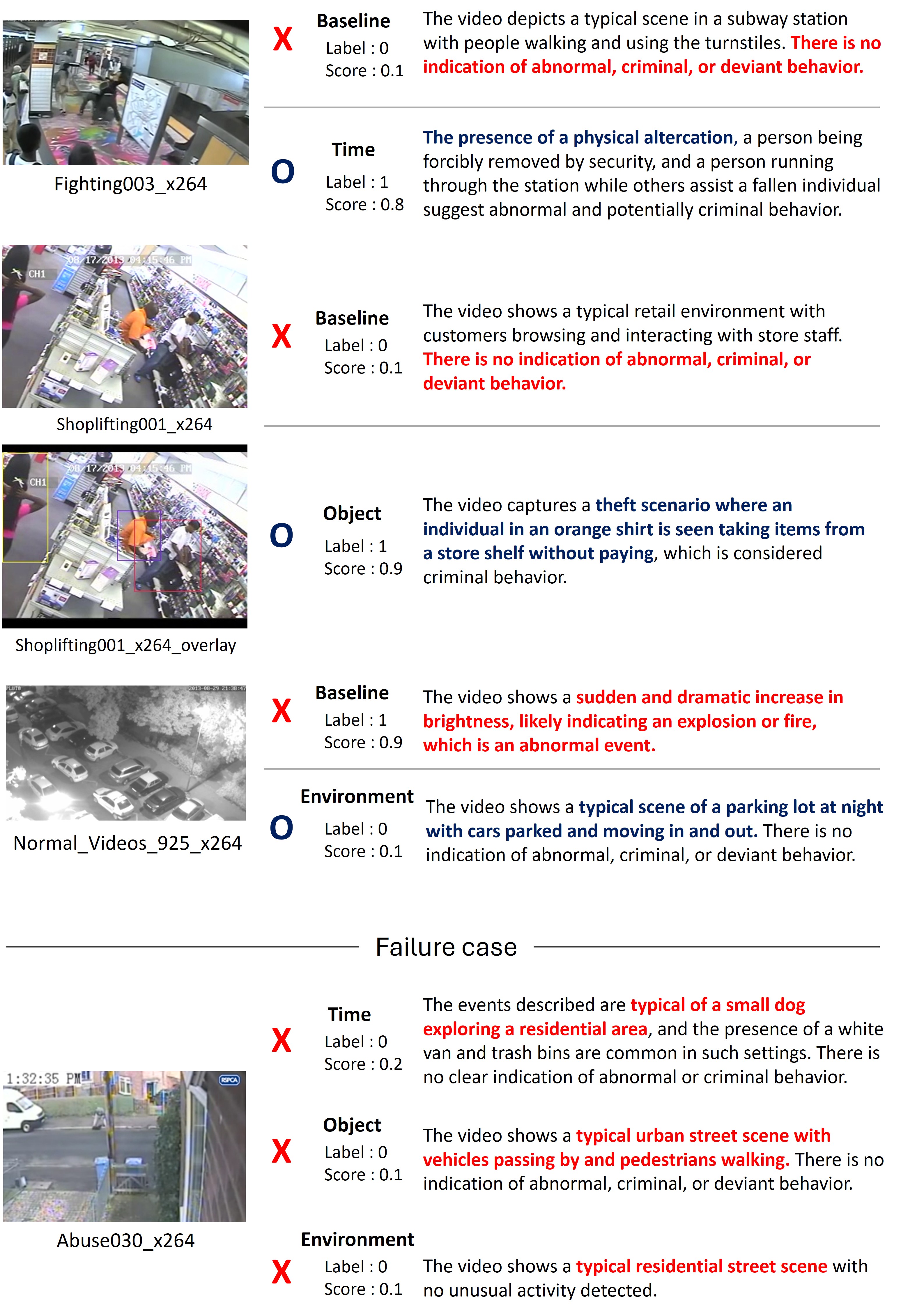}
    \caption{\textbf{Examples from UCF-Crime.} (Top to bottom) the first three examples show the advantage of our time, object, and environment context branches over the baseline, respectively, followed by a representative failure case.}
    \label{fig:qualitative_eval}
\end{figure}

To understand how the proposed context decomposition affects anomaly inference, we present qualitative examples from UCF-Crime in Fig.~\ref{fig:qualitative_eval}, including three success cases where a context-specific branch corrects the direct holistic baseline and one failure case where all context branches fail.

In the first example, the baseline interprets the subway scene as a typical public-space activity and fails to detect the anomaly. In contrast, our time branch captures the sequence of events, i.e. a physical fight and a person running while others assist a fallen individual, and correctly predicts an abnormal event. This example shows that time context is particularly effective when anomaly cues emerge from event progression.

In the second example, the baseline describes an ordinary retail scene and fails to detect the anomalous behavior. Our object branch focuses on the interaction between the person and the store shelf and correctly identifies the theft scenario. This shows that explicit object emphasis helps reveal fine-grained abnormal interactions that can be overlooked in holistic inference.

In the third example, the baseline incorrectly predicts an anomaly, triggered by a sudden increase in brightness. Our environment branch grounds its decision on the broader scene prior, i.e. a typical parking lot at night with parked and moving cars, and correctly classifies the video as normal. This case shows that the environment context can suppress spurious anomaly predictions caused by visually salient but non-anomalous changes.

The final example illustrates a representative failure case, where all three context branches misinterpret the scene as normal. This failure is caused by the uniform frame sampling strategy: the short anomalous moment is not captured in the sampled frames, and therefore the time, object, and environment branches all receive insufficient evidence for abnormality inference. This case shows a limitation of the current framework when brief abnormal events are missed during sparse sampling over time.

Overall, these examples show that the environment, object, and time branches provide complementary evidence for anomaly detection and can address different failure modes of holistic inference. The failure case also shows that our context-structured inference still depends on whether the sampled visual evidence contains the critical anomalous moment.

\subsection{Effect of sampling settings}
\label{subsec:sampling_results}

\begin{table}[t]
\caption{Video-level AUC-ROC (\%) under different uniform sampling budgets on UCF-Crime and UBnormal. Best results in each benchmark block are shown in bold, and the second-best results on UCF-Crime are underlined.}
\label{tab:auc_sampling}
\centering
\scriptsize
\setlength{\tabcolsep}{3.5pt}
\renewcommand{\arraystretch}{0.95}
\resizebox{\columnwidth}{!}{
\begin{tabular}{llcccc}
\toprule
\textbf{Dataset} & \textbf{\#Frames} & \textbf{Baseline} & \textbf{Ours (Max)} & \textbf{Ours (Mean)} & \textbf{Ours (Noisy-OR)} \\
\midrule
\multirow{4}{*}{UCF-Crime}
& 32  & 81.50 & 88.49 & \underline{89.04} & \textbf{89.08} \\
& 64  & 81.53 & 88.49 & \underline{90.70} & \textbf{90.73} \\
& 128 & 84.74 & 88.35 & \underline{91.69} & \textbf{91.76} \\
& 256 & 88.39 & 89.33 & \textbf{92.90} & \underline{92.88} \\
\midrule
\multirow{4}{*}{UBnormal}
& 32  & 59.19 & 70.52 & \textbf{72.16} & \textbf{72.16} \\
& 64  & 59.49 & 69.64 & \textbf{71.47} & \textbf{71.47} \\
& 128 & 59.81 & 69.42 & \textbf{71.27} & \textbf{71.27} \\
& 256 & 58.66 & 68.75 & \textbf{71.56} & \textbf{71.56} \\
\bottomrule
\end{tabular}
}
\end{table}

We further evaluate CSI-VAD under varying numbers of uniformly sampled frames on both UCF-Crime and UBnormal to examine whether the proposed context-structured inference remains effective under different visual input budgets. Table~\ref{tab:auc_sampling} reports the video-level AUC-ROC results of the baseline and the proposed score aggregation strategies with 32, 64, 128, and 256 sampled frames.

Across both benchmarks, CSI-VAD consistently outperforms the direct holistic baseline under all sampling settings. On UCF-Crime, all three aggregation strategies already improve over the baseline under sparse sampling, and the performance gap is maintained as the number of sampled frames increases. Among the three variants, Noisy-OR performs best at lower sampling budgets, while Mean Ensemble achieves the highest result at 256 frames.

A similar trend is observed on UBnormal, where CSI-VAD also consistently outperforms the baseline across all sampling settings. In particular, Mean Ensemble and Noisy-OR show the strongest and most stable performance, while the baseline remains substantially lower throughout. The proposed framework achieves its best UBnormal performance even under sparse sampling, indicating that context decomposition is effective even when the available visual evidence is limited.

Overall, Mean Ensemble and Noisy-OR are the most robust aggregation strategies across both benchmarks. These results show that decomposing anomaly inference into environment, object, and time contexts provides consistent gains over direct holistic inference, while remaining effective under both sparse and dense sampling budgets.

\subsection{Effect of binary classification}
\label{subsec:classification_results}

\begin{table}[t]
\caption{Video-level binary classification performance with 32 uniformly sampled frames on UCF-Crime and UBnormal. The rows $\lvert \mathcal{C} \rvert \ge 3$, $\lvert \mathcal{C} \rvert \ge 2$, and $\lvert \mathcal{C} \rvert \ge 1$ correspond to AND, majority, and OR aggregation, respectively.}
\label{tab:classification_results}
\centering
\scriptsize
\renewcommand{\arraystretch}{1.02}
\begin{tabular*}{\columnwidth}{@{\extracolsep{\fill}}llcccc@{}}
\toprule
\textbf{Dataset} & \textbf{Method} & \textbf{Accuracy} & \textbf{Precision} & \textbf{Recall} & \textbf{F1} \\
\midrule
\multirow{7}{*}{UCF-Crime}
& Baseline & 0.77 & \textbf{1.00} & 0.51 & 0.68 \\
& Obj ($y_o$) & 0.78 & 0.99 & 0.54 & 0.70 \\
& Env ($y_e$) & 0.76 & \textbf{1.00} & 0.51 & 0.67 \\
& Time ($y_t$) & 0.76 & 0.93 & 0.55 & 0.69 \\
& $\lvert \mathcal{C} \rvert \ge 3$ & 0.73 & \textbf{1.00} & 0.43 & 0.61 \\
& $\lvert \mathcal{C} \rvert \ge 2$ & 0.78 & \textbf{1.00} & 0.55 & 0.71 \\
& $\lvert \mathcal{C} \rvert \ge 1$ &
  \textbf{0.83} &
  0.93 &
  \textbf{0.69} &
  \textbf{0.80} \\
\midrule
\multirow{7}{*}{UBnormal}
& Baseline & 0.40 & 0.97 & 0.20 & 0.34 \\
& Obj ($y_o$) & 0.39 & \textbf{1.00} & 0.19 & 0.32 \\
& Env ($y_e$) & 0.43 & \textbf{1.00} & 0.23 & 0.38 \\
& Time ($y_t$) & 0.39 & \textbf{1.00} & 0.18 & 0.31 \\
& $\lvert \mathcal{C} \rvert \ge 3$ & 0.34 & \textbf{1.00} & 0.12 & 0.21 \\
& $\lvert \mathcal{C} \rvert \ge 2$ & 0.39 & \textbf{1.00} & 0.18 & 0.31 \\
& $\lvert \mathcal{C} \rvert \ge 1$ &
  \textbf{0.48} &
  \textbf{1.00} &
  \textbf{0.30} &
  \textbf{0.47} \\
\bottomrule
\end{tabular*}
\end{table}

We evaluate the final video-level binary decisions using 32 uniformly sampled frames, reporting Accuracy, Precision, Recall, and F1 score on both UCF-Crime and UBnormal. Table~\ref{tab:classification_results} compares the direct holistic baseline, the individual context-level labels, and the voting-based label aggregation rules. Here, $\lvert \mathcal{C} \rvert \ge 3$, $\lvert \mathcal{C} \rvert \ge 2$, and $\lvert \mathcal{C} \rvert \ge 1$ correspond to AND, majority, and OR aggregation, respectively.

Across both benchmarks, OR aggregation ($\lvert \mathcal{C} \rvert \ge 1$) achieves the strongest overall classification performance. Its main advantage comes from improving recall while maintaining high precision, leading to the best overall balance among the aggregation rules. A similar trend is observed on both UCF-Crime and UBnormal, indicating that permissive label aggregation is more effective than stricter voting rules for recovering abnormal videos.

These gains are mainly driven by improved recall, indicating that the proposed framework successfully recovers abnormal videos that are missed by single-context or holistic decisions. In contrast, the strict AND rule ($\lvert \mathcal{C} \rvert \ge 3$) results in the lowest recall on both benchmarks, showing that requiring anomaly predictions from all three contexts is overly conservative. Overall, the results support our claim that the three context branches provide complementary evidence and that permissive label aggregation is effective for capturing abnormal events.

To further analyze the gain of label aggregation, we examine abnormal category-level accuracy on UCF-Crime using OR aggregation ($\lvert \mathcal{C} \rvert \ge 1$). Table~\ref{tab:category_results} compares the baseline, the individual context-level decisions, and the final OR-based aggregation for each abnormal category.
The results show that OR aggregation improves performance in multiple categories while never degrading the baseline. In particular, large gains are observed for Burglary (+53.84), Assault (+33.34), Shoplifting (+23.81), Shooting (+21.74), Fighting (+20.00), Explosion (+14.29), and RoadAccidents (+13.04). Overall, OR aggregation improves 7 out of 13 abnormal categories and matches the baseline on the remaining 6 categories.

These improvements support the interpretation that the three context branches provide complementary evidence for anomaly detection. While a single branch may capture only part of the abnormal cue, aggregating the context-level labels allows CSI-VAD to recover anomalies missed by the holistic baseline. This category-wise trend is consistent with the recall gains observed in Table~\ref{tab:classification_results}, showing that context decomposition helps detect a wider range of abnormal events.

\newcommand{\gain}[1]{{\tiny (#1)}}
\newcommand{\nogain}{{\tiny (+0)}}
\begin{table}[t]
\caption{Abnormal category-level accuracy (\%) on UCF-Crime using OR aggregation ($\lvert \mathcal{C} \rvert \ge 1$). In the last column, the parenthesized value following the OR aggregation result indicates the absolute improvement over the baseline.}
\label{tab:category_results}
\centering
\footnotesize
\setlength{\tabcolsep}{2pt}
\renewcommand{\arraystretch}{1.10}
\resizebox{0.9\columnwidth}{!}{
\begin{tabular}{lccccl}
\toprule
\textbf{Category} & \textbf{Baseline} & \textbf{Env ($y_e$)} & \textbf{Obj ($y_o$)} & \textbf{Time ($y_t$)} & \textbf{$\lvert \mathcal{C} \rvert \ge 1$} \\
\midrule
Abuse         & 0.00  & 0.00  & 0.00  & 0.00  & 0.00 \nogain \\
Arrest        & 80.00 & 80.00 & 80.00 & 80.00 & 80.00 \nogain \\
Arson         & 88.89 & 88.89 & 77.78 & 66.67 & 88.89 \nogain \\
Assault       & 33.33 & 33.33 & 66.67 & 33.33 & \textbf{66.67 \gain{+33.34}} \\
Burglary      & 23.08 & 30.77 & 53.85 & 61.54 & \textbf{76.92 \gain{+53.84}} \\
Explosion     & 76.19 & 66.67 & 85.71 & 71.43 & \textbf{90.48 \gain{+14.29}} \\
Fighting      & 80.00 & 80.00 & 80.00 & 80.00 & \textbf{100.00 \gain{+20.00}} \\
RoadAccidents & 43.48 & 43.48 & 43.48 & 47.83 & \textbf{56.52 \gain{+13.04}} \\
Robbery       & 60.00 & 60.00 & 40.00 & 60.00 & 60.00 \nogain \\
Shooting      & 52.17 & 60.87 & 47.83 & 65.22 & \textbf{73.91 \gain{+21.74}} \\
Shoplifting   & 19.05 & 9.52  & 23.81 & 19.05 & \textbf{42.86 \gain{+23.81}} \\
Stealing      & 60.00 & 60.00 & 40.00 & 60.00 & 60.00 \nogain \\
Vandalism     & 80.00 & 80.00 & 80.00 & 60.00 & 80.00 \nogain \\
\bottomrule
\end{tabular}
}
\end{table}

\section{Conclusion}
\label{sec:conclusion}

We proposed CSI-VAD, a training-free video anomaly detection framework that replaces holistic LVLM inference with context-structured inference over environment, object, and time cues.
By decomposing anomaly detection into complementary branches and aggregating their outputs, CSI-VAD reduces inference complexity and improves video-level detection. Experiments on UCF-Crime and UBnormal showed consistent gains over the direct holistic baseline and competitive performance against existing methods.
The failure case analysis shows that sparse uniform sampling can miss brief anomalous moments, leaving insufficient evidence for all branches. In addition, CSI-VAD requires multiple LVLM/LLM calls along with object detection and tracking, introducing non-negligible inference cost for real-time surveillance deployment. Future work will therefore extend CSI-VAD toward finer-grained detection over time or at the frame level, as well as more efficient inference through lightweight backbones or adaptive context selection.

\section*{Acknowledgment}
This work was supported by the Korea Institute for Advancement of Technology (KIAT) grant funded by the Korea Government (MOTIE) (P0028485, GITCC) and the BK21 FOUR (Fostering Outstanding Universities for Research) funded by the Ministry of Education (MOE, Korea) and National Research Foundation of Korea (NRF). This research utilised Queen Mary's Apocrita HPC facility, supported by QMUL Research-IT (\url{http://doi.org/10.5281/zenodo.438045}).

\bibliographystyle{IEEEtran}
\bibliography{references}

@inproceedings{sultani2018real,
  author    = {Sultani, Waqas and Chen, Chen and Shah, Mubarak},
  title     = {Real-world anomaly detection in surveillance videos},
  booktitle = {IEEE Conf. Comput. Vis. Pattern Recognit.},
  year      = {2018}
}

@inproceedings{tian2021weakly,
  author    = {Tian, Yu and Pang, Guansong and Chen, Yuanhong and Singh, Rajvinder and Verjans, Johan W and others},
  title     = {Weakly-supervised video anomaly detection with robust temporal feature magnitude learning},
  booktitle = {IEEE/CVF Int. Conf. Comput. Vis.},
  year      = {2021}
}

@inproceedings{chen2023tevad,
  author    = {Chen, Weiling and Ma, Keng Teck and Yew, Zi Jian and Hur, Minhoe and Khoo, David Aik-Aun},
  title     = {{TEVAD}: Improved video anomaly detection with captions},
  booktitle = {IEEE/CVF Conf. Comput. Vis. Pattern Recognit.},
  year      = {2023}
}

@inproceedings{zanella2024harnessing,
  author    = {Zanella, Luca and Menapace, Willi and Mancini, Massimiliano and Wang, Yiming and Ricci, Elisa},
  title     = {Harnessing large language models for training-free video anomaly detection},
  booktitle = {IEEE/CVF Conf. Comput. Vis. Pattern Recognit.},
  year      = {2024}
}

@inproceedings{alayrac2022flamingo,
  author    = {Alayrac, Jean-Baptiste and Donahue, Jeff and Luc, Pauline and Miech, Antoine and Barr, Iain and others},
  title     = {Flamingo: A visual language model for few-shot learning},
  booktitle = {Adv. Neural Inf. Process. Syst.},
  year      = {2022}
}

@inproceedings{li2023blip2,
  author    = {Li, Junnan and Li, Dongxu and Savarese, Silvio and Hoi, Steven C. H.},
  title     = {{BLIP-2}: Bootstrapping language-image pre-training with frozen image encoders and large language models},
  booktitle = {Int. Conf. Mach. Learn.},
  year      = {2023}
}

@article{pang2021deep,
  author    = {Pang, Guansong and Shen, Chunhua and Cao, Longbing and Van Den Hengel, Anton},
  title     = {Deep learning for anomaly detection: A review},
  journal   = {ACM Comput. Surv.},
  volume    = {54},
  number    = {2},
  pages     = {1--38},
  year      = {2021}
}

@inproceedings{yang2024context,
  author    = {Yang, Zhengye and Radke, Richard J},
  title     = {Context-aware video anomaly detection in long-term datasets},
  booktitle = {IEEE/CVF Conf. Comput. Vis. Pattern Recognit.},
  year      = {2024}
}

@inproceedings{acsintoae2022ubnormal,
  author    = {Acsintoae, Andra and Florescu, Andrei and Georgescu, Mariana-Iuliana and Mare, Tudor and Sumedrea, Paul and others},
  title     = {{UBnormal}: New benchmark for supervised open-set video anomaly detection},
  booktitle = {IEEE/CVF Conf. Comput. Vis. Pattern Recognit.},
  year      = {2022}
}

@inproceedings{wu2022self,
  author    = {Wu, Jhih-Ciang and Hsieh, He-Yen and Chen, Ding-Jie and Fuh, Chiou-Shann and Liu, Tyng-Luh},
  title     = {Self-supervised sparse representation for video anomaly detection},
  booktitle = {Eur. Conf. Comput. Vis.},
  year      = {2022}
}

@inproceedings{chen2023mgfn,
  author    = {Chen, Yingxian and Liu, Zhengzhe and Zhang, Baoheng and Fok, Wilton and Qi, Xiaojuan and others},
  title     = {{MGFN}: Magnitude-contrastive glance-and-focus network for weakly-supervised video anomaly detection},
  booktitle = {AAAI Conf. Artif. Intell.},
  year      = {2023}
}

@article{pu2024learning,
  author    = {Pu, Yujiang and Wu, Xiaoyu and Yang, Lulu and Wang, Shengjin},
  title     = {Learning prompt-enhanced context features for weakly-supervised video anomaly detection},
  journal   = {IEEE Trans. Image Process.},
  volume    = {33},
  pages     = {4923--4936},
  year      = {2024}
}

@article{zhou2024batchnorm,
  author    = {Zhou, Yixuan and Qu, Yi and Xu, Xing and Shen, Fumin and Song, Jingkuan and others},
  title     = {Batchnorm-based weakly supervised video anomaly detection},
  journal   = {IEEE Trans. Circuits Syst. Video Technol.},
  volume    = {34},
  number    = {12},
  pages     = {13642--13654},
  year      = {2024}
}

@inproceedings{gao2025stead,
  author    = {Gao, Andrew and Liu, Jun},
  title     = {{STEAD}: Spatio-temporal efficient anomaly detection for time and compute sensitive applications},
  booktitle = {IEEE/RSJ Int. Conf. Intell. Robots Syst.},
  year      = {2025}
}

@inproceedings{jung2025dual,
  author    = {Jung, Seoik and Song, Taekyung and Daniel, Joshua Jordan and Lee, JinYoung and Lee, SungJun},
  title     = {{DUAL-VAD}: Dual Benchmarks and Anomaly-Focused Sampling for Video Anomaly Detection},
  booktitle = {IEEE/IEIE Int. Conf. Consum. Electron.-Asia},
  year      = {2025}
}

@inproceedings{ye2025vera,
  author    = {Ye, Muchao and Liu, Weiyang and He, Pan},
  title     = {{VERA}: Explainable video anomaly detection via verbalized learning of vision-language models},
  booktitle = {IEEE/CVF Conf. Comput. Vis. Pattern Recognit.},
  year      = {2025}
}

@article{Qwen2.5-VL,
  author    = {Bai, Shuai and Chen, Keqin and Liu, Xuejing and Wang, Jialin and Ge, Wenbin and others},
  title     = {{Qwen2.5-VL} Technical Report},
  journal   = {arXiv preprint arXiv:2502.13923},
  year      = {2025}
}

@article{qwen2.5,
  author    = {Yang, An and Yang, Baosong and Zhang, Beichen and Hui, Binyuan and Zheng, Bo and others},
  title     = {{Qwen2.5} Technical Report},
  journal   = {arXiv preprint arXiv:2412.15115},
  year      = {2024}
}

@article{du2023strongsort,
  author    = {Du, Yunhao and Zhao, Zhicheng and Song, Yang and Zhao, Yanyun and Su, Fei and others},
  title     = {{StrongSORT}: Make {DeepSORT} great again},
  journal   = {IEEE Trans. Multimedia},
  volume    = {25},
  pages     = {8725--8737},
  year      = {2023}
}

@article{robinson2025rf,
  author    = {Robinson, Isaac and Robicheaux, Peter and Popov, Matvei and Ramanan, Deva and Peri, Neehar},
  title     = {{RF-DETR}: Neural architecture search for real-time detection transformers},
  journal   = {arXiv preprint arXiv:2511.09554},
  year      = {2025}
}

@article{borodin2025benchmarking,
  author  = {Borodin, Kirill and Kondrashov, Kirill and Vasiliev, Nikita and Gladkova, Ksenia and Larina, Inna and others},
  title   = {Benchmarking Compact {VLMs} for Clip-Level Surveillance Anomaly Detection Under Weak Supervision},
  journal = {J. Imaging},
  volume  = {11},
  number  = {11},
  pages   = {400},
  year    = {2025}
}

@book{pearl1988probabilistic,
  author    = {Pearl, Judea},
  title     = {Probabilistic Reasoning in Intelligent Systems: Networks of Plausible Inference},
  publisher = {Morgan Kaufmann},
  year      = {1988}
}

@article{kittler1998combining,
  author  = {Kittler, Josef and Hatef, Mohamad and Duin, Robert P. W. and Matas, Jiri},
  title   = {On Combining Classifiers},
  journal = {IEEE Transactions on Pattern Analysis and Machine Intelligence},
  volume  = {20},
  number  = {3},
  pages   = {226--239},
  year    = {1998}
}

@article{lam1997majority,
  author  = {Lam, Louisa and Suen, Ching Y.},
  title   = {Application of Majority Voting to Pattern Recognition: An Analysis of Its Behavior and Performance},
  journal = {IEEE Transactions on Systems, Man, and Cybernetics - Part A: Systems and Humans},
  volume  = {27},
  number  = {5},
  pages   = {553--568},
  year    = {1997}
}

\end{document}